# Image Restoration in Non-Linear Filtering Domain using MDB approach

S. K. Satpathy, S. Panda, K. K. Nagwanshi* and C. Ardil

*Abstract*—This paper proposes a new technique based on non-linear Minmax Detector Based (MDB) filter for image restoration. The aim of image enhancement is to reconstruct the true image from the corrupted image. The process of image acquisition frequently leads to degradation and the quality of the digitized image becomes inferior to the original image. Image degradation can be due to the addition of different types of noise in the original image. Image noise can be modeled of many types and impulse noise is one of them. Impulse noise generates pixels with gray value not consistent with their local neighborhood. It appears as a sprinkle of both light and dark or only light spots in the image. Filtering is a technique for enhancing the image. Linear filter is the filtering in which the value of an output pixel is a linear combination of neighborhood values, which can produce blur in the image. Thus a variety of smoothing techniques have been developed that are non linear. Median filter is the one of the most popular non-linear filter. When considering a small neighborhood it is highly efficient but for large window and in case of high noise it gives rise to more blurring to image. The Centre Weighted Mean (CWM) filter has got a better average performance over the median filter. However the original pixel corrupted and noise reduction is substantial under high noise condition. Hence this technique has also blurring affect on the image. To illustrate the superiority of the proposed approach, the proposed new scheme has been simulated along with the standard ones and various restored performance measures have been compared.

## I. INTRODUCTION

IMAGE processing algorithms are designed to handle different problem domains. Some of the broad area includes (i) Image Representation and Modeling Image Transforms, (ii) Image Enhancement Image Restoration, (iii) Image Analysis Image Reconstruction, and (iv) Image Data Compression. Many image-processing algorithms are designed to deal with the above-mentioned problems. Image restoration is principally refers to the removal or minimization of known

___________________________________________

S. K. Satpathy is with the Department of Computer Science & Engineering, Rungta College of Engineering & Technology, Bhilai CG 490023 India (Phone: 922-935-5518; fax: 788-228-6480; e-mail: sks_sarita@yahoo.com).

S. Panda is with the National Institute of Science and Technology (NIST) Berhampur, OR 760007 India (e-mail: panda_sidhartha@rediffmail.com).

K. K. Nagwanshi is with the Department of Computer Science & Engineering, Rungta College of Engineering & Technology, Bhilai CG 490023 India (e-mail: kapilkn@gmail.com).

C. Ardil is with National Academy of Aviation, AZ1045, Baku, Azerbaijan, Bina, 25th km, NAA (e-mail: cemalardil@gmail.com).

degradation in an image *i.e.* the recovery of an image that has been degraded by some prior knowledge of the degradation phenomenon. Consequently the restoration techniques are oriented towards modeling the degradation and applying the inverse process in order to recover the original image. Instead of using traditional frequency domain concept, here some algebraic methods have been used for the system, which has the advantage of computational complexity over the frequency domain restoration technique. When one would like to remove the noise, it follows certain filtering operation where the signal has to be passed through a filter and the filter in turn removes the undesirable components. Frequently linear techniques are used because linear filters are easy to implement and design. Further they are optimal among the classes of all filtering operations when the noise is additive and Gaussian [1].

*A. Image Noise Model*

When it has been discussed on noise it can be getting introduced in the image, either at the time of image generation (e.g. when we use camera and photographic films to capture an image) or at the time of image transmission. Accordingly different categories of noise having certain characteristics. In photographic films; the recording noise is mainly due to the silver grains that precipitate during the film exposure. They behave randomly during both film exposure and development. They are also randomly located on the films. This kind of noise, which is due to silver grains, is called film grain noise. This is a Poisson process and becomes a Gaussian process in its limit. The film grain noise does not any statistical correlation for distance between samples greater than the grain size. Therefore film grain noise is a white noise two-dimensional random process. In photo electronic detectors, two kind of noise appears: (a) Thermal Noise: Its source is the various electronic circuits. It's a two-dimensional additive white zero-mean Gaussian noise, and (b) Photoelectron Noise: it is produced by random fluctuation of the number of photons on the light sensitive surface of the detector [2]. If its level is low it can be treated as a poison distributed noise otherwise it can be treated as a Gaussian distributed noise. Thus it can be said as signal dependent noise. Another kind of noise that is present during the image transmission is Salt-Pepper noise [3]. It appears as black and/or white impulse of the image. Its source is usually man-made or atmospheric noise, which appears as impulsive noise. It has following form: where $z(k,j)$ denotes the impulse and $i(k,j)$ denotes the original image intensity at the pixel *(k, j)*. In case of the-CCD cameras, the main form of noise is the transfer loss noise.

*B. Digital Filters*

Elimination of noise is one of the major works to be done in computer vision and image processing, as noise leads to the error in the image. Presence of noise is manifested by undesirable information, which is not at all related to the image under study, but in turn disturbs the information present in the image. It is translated into values, which are getting added or subtracted to the true gray level values on a gray level pixel. These unwanted noise information can be introduced because of so many reasons like: acquisition process due to cameras quality and restoration, acquisition condition, such as illumination level, calibration and positioning or it can be a function of the scene environment. Noise elimination is a main concern in computer vision and image processing. A digital filter [1][3] is used to remove noise from the degraded image. As any noise in the image can be result in serious errors. Noise is an unwanted signal, which is manifested by undesirable information. Thus the image, which gets contaminated by the noise, is the degraded image and using different filters can filter this noise. Thus filter is an important subsystem of any signal processing systems. Thus filters are used for image enhancement, as it removes undesirable signal components from the signal of interest. Filters are of different type i.e. linear filters or nonlinear filters [20]. In early times, as the signals handled were analog, filters used are of analog. Gradually digital filters were took over the analog systems because of their flexibility, low cost, programmability, reliability, etc. for these reasons digital filters are designed which works with digital signals. The design-of digital filters involves three basic steps: (i) the specification of the desired properties of the system,(ii) the approximation of these specifications using a causal discrete-time system, and (iii) the realization of the system using finite precision arithmetic.

The paper has been organized in the following manner; section II proposes problem formulation and solution methodology, section III describes the result and discussion, section IV gives concluding remarks and further works and finally section V incorporates all the references been made for completion of this work.

II. PROBLEM FORMULATION AND SOLUTION METHODOLOGY

As it had seen that noise elimination is a main concern in computer vision and image processing. For example in many applications where operators based on computing image derivatives is applied, any noise in the image can results in serious errors. Noise presence is manifested by undesirable information, not related to the scene under study, which perturbs the information relative to the form observable in the image. It is translated to more or less severe values, which are added or subtracted to the original values on a number of pixels. Noise is of many types. Thus image noise can be Gaussian, Uniform or impulsive distribution. Here we will discuss about the, impulse noise. Impulse noise generates pixels with gray level values not consistent with their local neighborhood values. The impulse noise appears as sprinkle of light or dark spots in the image. This impulse noise can be eliminated or the degraded image can be enhanced by the use of advance filter. Due to certain disadvantages of linear filters, nonlinear method of filtering has been proposed in this paper. The filter cannot be modeled by convolutions fall into the category of nonlinear filter. Nonlinear filter can be very effective in removing the impulse noise. The median filter replaces the middle pixel value with the median value of the window. The most popular non-linear filter is the median filter. When considering a small neighborhood it is highly efficient and has proved to be very effective in removing noise of an impulsive nature despite its simple definition. Nevertheless, the median filter often fails to provide sufficient smoothing of non-impulsive noise and its result is sometimes unpredictable. Several techniques have been proposed which try to take the advantage of the average performance of the median filter, either to evaluate noise density, set up parameters or to guide the filtering process.

The weighted median (WM) filter [9] is an extension of the median filter, which -gives more weight to some values within the window. This WM filter allows a degree of control of the smoothing behavior through the weights that can be set, and therefore, it is a promising image enhancement technique. A special case of WM filter called center weighted median (CWM) filter [10]. This filter gives more weight only to the central value of a window, and thus it is easier to design and implement than general WM filters. These approaches involve a preliminary identification of corrupted pixels in an effort to prevent alteration of true pixel values.

A nonlinear smoothing filter namely the Peak-and-Valley filter is proposed to reduce impulsive-like noise while modifying the gray levels of the image as little as possible, resulting in a maximum preservation of the original information [11]. It could be used as a basis for the design of more sophisticated impulsive noise elimination filters, replacing the conventional median filter in performing the preliminary processing. In this paper thrust has been made to devise a filtering scheme to remove impulse noise from images such that the scheme should work at high noise conditions and should perform superior to the existing schemes in terms of noise rejection and retention of original image properties. The detection scheme is devised keeping the CWM filter in mind whereas the median filter is used for the filtering operation for detected noisy pixels. Extensive simulation has been carried out to compare the performance of the proposed filter with other standard schemes.

*A. Solution Methodology*

In any noise removal schemes, attention is given to remove noise from images in addition to keeping as much original properties as possible. However, since both the objectives are contradicting in nature, it is not possible for any nonlinear scheme to fulfill both the objectives. It has been observed through the simulation of the existing schemes [10,11,20] that they also fail in providing satisfactory results under high noise conditions. Since impulse noise is not uniformly distributed across the image, it is desirable to replace the corrupted ones through a suitable filter. For this purpose, a preprocessing is required to detect the corrupted location prior to filtering.

### B. Mathematical Analysis

To assess the performance of the proposed filters for removal of impulse noise and to evaluate their comparative performance, different standard performance indices have been used in the thesis. These are defined as follows:

**Peak Signal to Noise Ratio (PSNR):** It is measured in decibel (dB) and for gray scale image it is defined as:

$$\text{PSNR (dB)} = 10\log_{10}\left[\frac{\sum_i \sum_j 255^2}{\sum_i \sum_j (S_{i,j} - \hat{S}_{i,j})^2}\right] \quad (1)$$

Where $S_{i,j}$ and $\hat{S}_{i,j}$ are the original and restored image pixels respectively. The higher the PSNR in the restored image, the better is its quality.

**Signal to Noise Ratio Improvement (SNRI):** SNRI in dB is defined as the difference between the Signal to Noise Ratio (SNR) of the restored image in dB and SNR of restored image in dB i.e.

SNRI (dB) =
SNR of restored image in dB- SNR of noisy image in dB   (2)

Where,

SNR of restored image dB=

$$10\log_{10}\left[\frac{\sum_i \sum_j S_{i,j}^2}{\sum_i \sum_j (S_{i,j} - \hat{S}_{i,j})^2}\right] \quad (3)$$

SNR of Noisy image in dB =

$$10\log_{10}\left[\frac{\sum_i \sum_j S_{i,j}^2}{\sum_i \sum_j (S_{i,j} - X_{i,j})^2}\right] \quad (4)$$

Where, $X_{i,j}$ is Noisy image pixel

The higher value of SNRI reflects the better visual and restoration performance.

**Percentage of Spoiled Pixels (POSP):** It may be defined as the number of unaffected original pixels replaced with a different gray value after filtering *i.e.*

POSP =

$$\left[\frac{\text{Number of original pixels change their grayscale}}{\text{Number of non - noisy pixels}}\right] \times 100 \quad (5)$$

**Percentage of Noise Attenuated (PONA):** It may be defined as the number of pixels getting improved after being filtered.

PONA =

$$\left[\frac{\text{Number of noisy pixels getting improved}}{\text{Total number of noisy pixels}}\right] \times 100 \quad (6)$$

This parameter reflects the capability of the impulse noise detector used prior to filtering. Small value of POSP shows the improved performance of the detector. High value of POSP leads to removal of original image properties and. edge jitters in restored images. The more the percentage the better will be the attenuation characteristic of the filter filtering.

### C. Proposed Algorithms

| *Algorithm: MDB Filter* |  |
|---|---|
| Step 1. | Take corrupted image (X). |
| Step 2. | Take a 3 x 3 window (W). Let the center pixel be the test pixel. |
| Step 3. | Shift the window row wise then column wise to cover the entire pixel in the image and repeat Step4 and Step5. |
| Step 4. | If (test pixel < min (rest of the pixel in W) OR ( test pixel > max ( rest of the pixel in W) then the test pixel is corrupted. |
| Step 5. | If the test pixel is corrupted apply median filter to the test filter in the window W. |
| Step 6. | Stop. |

### III. RESULT AND DISCUSSION

The quantitative results has been given in table [Table I – Table VI] for the standard LENA image, for different percentage of noise, starting from 5% to 30% with a step of 5%, and the comparative analysis has been presented in figure [fig. 3 and fig. 4] for both LENA and CAMMAN image showing the performance of CWM filters for gain =1 and 2 respectively from 5% to 30% along with Salt and Pepper noise. Along with the figures and tables some graphs, shown in figure [graph1 – graph 2], has also been given, for all the quantitative measures, for both LENNA and CAMMAN image to have a quick insight into the comparative performance of the existing filters along with the proposed one i.e. MDB filter.

TABLE I
% NOISE ATTENUATED FOR LENA.TIF

| % Impulse Noise | CWM for K=1 | CWM for K=2 | MDB |
|---|---|---|---|
| 5 | 98.7562 | 97.6801 | 99.757 |
| 10 | 98.8724 | 96.4177 | 99.8234 |
| 15 | 97.572 | 92.7032 | 99.7087 |
| 20 | 97.9097 | 85.4656 | 99.5008 |
| 25 | 96.1055 | 80.0837 | 99.2512 |
| 30 | 95.5977 | 75.0516 | 99.2028 |

TABLE II
% IMAGE SPOIL FOR LENA.TIF

| % Impulse Noise | CWM for K=1 | CWM for K=2 | MDB |
|---|---|---|---|
| 5 | 47.3299 | 27.2792 | 21.6439 |
| 10 | 46.5746 | 25.7312 | 19.4961 |
| 15 | 46.5171 | 23.633 | 18.0531 |
| 20 | 46.1974 | 22.6867 | 16.9588 |
| 25 | 44.5388 | 20.9869 | 14.7011 |
| 30 | 44.4892 | 20.2045 | 13.566 |

TABLE III
% SNR RESTORED FOR LENA.TIF

| % Impulse Noise | CWM for K=1 | CWM for K=2 | MDB |
|---|---|---|---|
| 5 | 21.9903 | 23.3947 | 22.9579 |
| 10 | 21.101 | 20.5031 | 22.2465 |
| 15 | 19.4544 | 17.3634 | 21.8329 |
| 20 | 18.5534 | 14.5687 | 20.7446 |
| 25 | 16.7922 | 13.0016 | 19.4695 |
| 30 | 15.9019 | 11.3396 | 18.8912 |

TABLE IV
% SNR OF NOISY FOR LENA.TIF

| % Impulse Noise | CWM for K=1 | CWM for K=2 | MDB |
|---|---|---|---|
| 5 | 13.5613 | 13.6655 | 13.4663 |
| 10 | 10.5288 | 10.4552 | 10.3947 |
| 15 | 8.9576 | 8.872 | 8.9921 |
| 20 | 7.6403 | 7.5795 | 7.6698 |
| 25 | 6.5894 | 6.7567 | 6.5552 |
| 30 | 5.8324 | 5.941 | 5.926 |

TABLE V
%DIFFERENCE IN SNR FOR LENA.TIF

| % Impulse Noise | CWM for K=1 | CWM for K=2 | MDB |
|---|---|---|---|
| 5 | 8.4291 | 9.7291 | 9.4916 |
| 10 | 10.5721 | 10.0479 | 11.8519 |
| 15 | 10.4968 | 8.4915 | 12.8408 |
| 20 | 10.1931 | 6.9892 | 13.0949 |
| 25 | 10.2028 | 6.2449 | 12.9143 |
| 30 | 10.0695 | 5.3986 | 12.9652 |

TABLE VI
PEAK SIGNAL TO NOISE RATIO FOR LENA.TIF

| % Impulse Noise | CWM for K=1 | CWM for K=2 | MDB |
|---|---|---|---|
| 5 | 27.4687 | 28.8749 | 28.4385 |
| 10 | 26.5805 | 25.9823 | 27.7254 |
| 15 | 24.9343 | 22.8439 | 27.312 |
| 20 | 24.0316 | 20.0496 | 26.2449 |
| 25 | 22.2725 | 18.4822 | 24.9492 |
| 30 | 21.3819 | 16.8204 | 24.3714 |

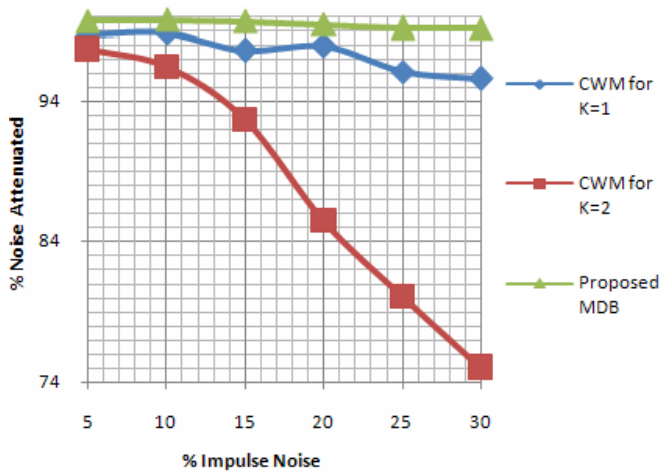

Fig. 1: Comparison of % Impulse Noise vs. % Noise Attenuated

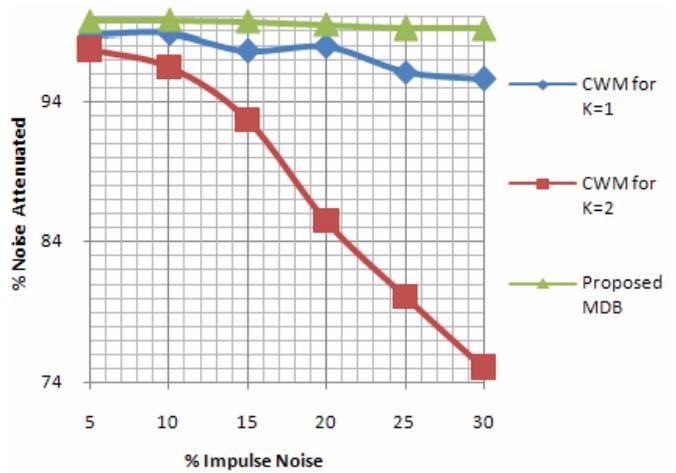

Fig. 2: Comparison of % Impulse Noise vs. PSNR

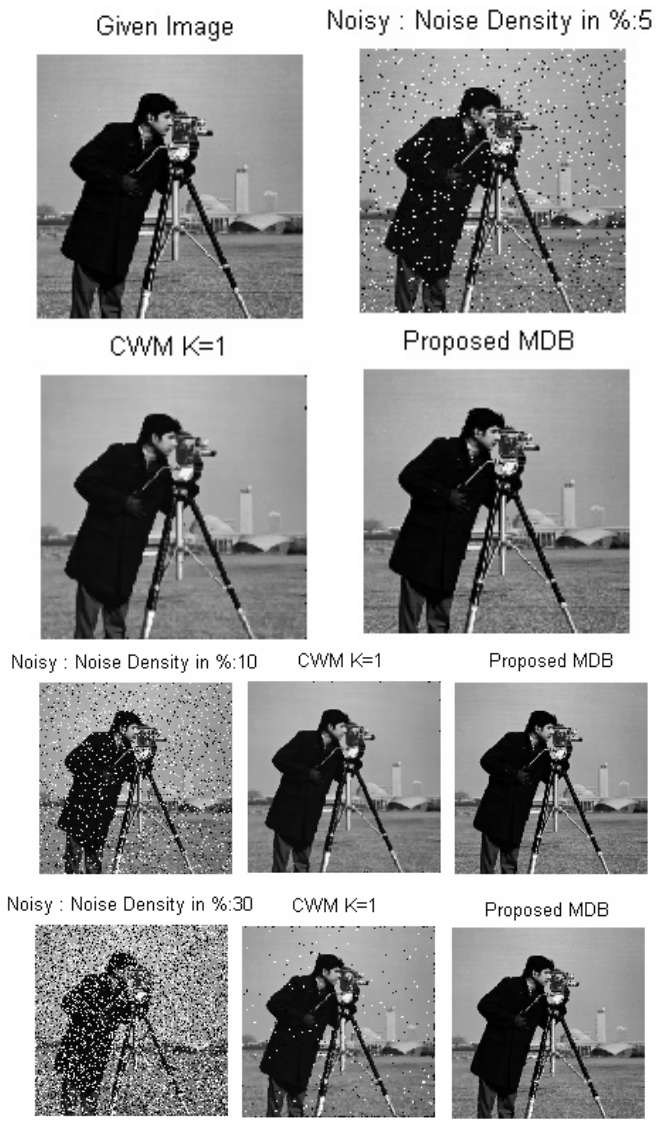

Fig. 3: Result with Cameraman.tif 256x256 image at various noise levels

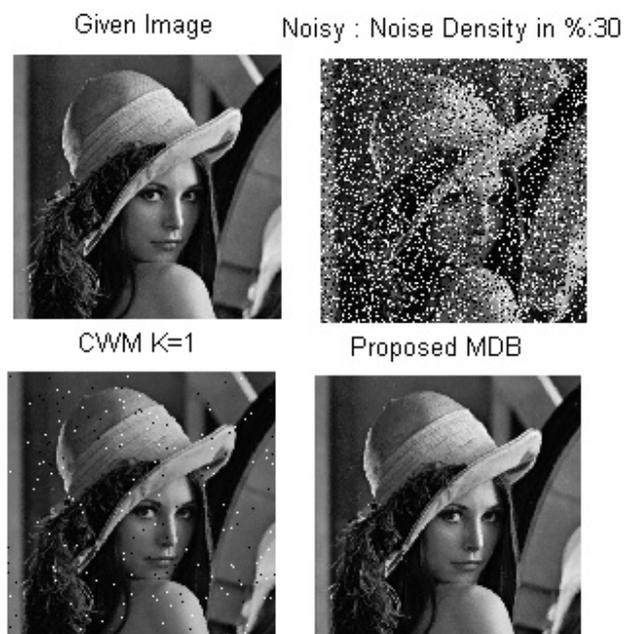

Fig. 3: Result with Lena.tif 256x256 image at 30%Impulse Noise

## IV. CONCLUSION AND FURTHER ENHANCEMENT

This paper proposed new non-linear filters to remove the impulse noise from the images. To illustrate the efficiency of the proposed MDB schemes, we have simulated the new schemes along with the existing ones and various restored measures have been compared. All the filtering techniques have been simulated in MATLAB 7.1 with Pentium-IV processor. The schemes are simulated using standard images LENA and CAMMAN. The impure contaminations include Salt and Pepper noise. The proposed schemes MDB filter is found to be superior, i.e. better restored results and other parameter for restoration compared to the existing schemes when Salt and Pepper impulse noise is considered. The noise removal schemes can be extended based on other types of detection like Fuzzy and ANN. There is another emerging area of image restoration technique, called Blind De-Convolution technique. In this direction further work can be extended by utilizing soft computing techniques, like FUZZY and ANN, so that the properties of images can be better retained in the restored images.

ACKNOWLEDGMENT

The authors wish thanks to Dr. A. Jagadeesh, Director, RCET-Bhilai for his kind support. We would also especially grateful to Mr. Surabh Rungta for providing necessary facilities to incorporate this research work.

**Susanta K. Satpathy** (M'08) is working as a professor at Rungta College of Engineering and Technology,Bhilai, Chhattisgarh. He received the M. Tech degree in Computer Science and Engineering from NIT, Rourkela in 2003 and B.Tech. dgree in Computer Science and Engineering in 1995. He has published above 10 papers in various Journals and conferences. He is life time member of CSI and ISTE since 2008. He worked in various other engineering colleges for about 14 years. His area of reasearch includes Signal processing, image processing and information system and security.

**Sidhartha Panda** is working as a Professor at National Institute of Science and Technology (NIST), Berhampur, Orissa, India. He received the Ph.D. degree from Indian Institute of Technology, Roorkee, India in 2008, M.E. degree in Power Systems Engineering from UCE, Burla in 2001 and B.E. degree in Electrical Engineering in 1991. Earlier, he worked in various engineering colleges for about 15 years. He has published above 50 papers in various International Journals and acting as a reviewer for some IEEE and Elsevier Journals. His biography has been included in Marquis' "Who's Who in the World" USA, for 2010 edition. His areas of research include, power system dynamic stability, FACTS, optimization techniques, model order reduction, distributed generation, image processing and wind energy.

**Kapil K. Nagwanshi** was born in August 1978 in Chhindwara district, Madhya Pradesh, India. He is graduated from GG University, Bilaspur, a Central University of Chhattisgarh State, India, in Computer Science & Engineering in the year 2001 and later did his post graduation in Computer Technology & Application from Chhattisgarh Swami Vivekanand Technical University, Bhilai, India. Currently he is working as a Reader in RCET Bhilai, and he has published more than 28 research papers in reputed journals, national and international conferences. His research area includes, signal processing, and image processing, and information systems and security.

**C. Ardil** is with National Academy of Aviation, AZ1045, Baku, Azerbaijan, Bina, 25th km, NAA